\PassOptionsToPackage{unicode}{hyperref}
\PassOptionsToPackage{hyphens}{url}
\PassOptionsToPackage{dvipsnames,svgnames,x11names}{xcolor}
\documentclass[
  10pt,
]{article}
\usepackage{xcolor}
\usepackage[margin=0.85in]{geometry}
\usepackage{amsmath,amssymb}
\usepackage{iftex}
\ifPDFTeX
  \usepackage[T1]{fontenc}
  \usepackage[utf8]{inputenc}
  \usepackage{textcomp} 
\else 
  \usepackage{unicode-math} 
  \defaultfontfeatures{Scale=MatchLowercase}
  \defaultfontfeatures[\rmfamily]{Ligatures=TeX,Scale=1}
\fi
\usepackage{lmodern}
\ifPDFTeX\else
\fi
\IfFileExists{upquote.sty}{\usepackage{upquote}}{}
\IfFileExists{microtype.sty}{
  \usepackage[]{microtype}
  \UseMicrotypeSet[protrusion]{basicmath} 
}{}
\makeatletter
\@ifundefined{KOMAClassName}{
  \IfFileExists{parskip.sty}{%
    \usepackage{parskip}
  }{
    \setlength{\parindent}{0pt}
    \setlength{\parskip}{6pt plus 2pt minus 1pt}}
}{
  \KOMAoptions{parskip=half}}
\makeatother
\usepackage{longtable,booktabs,array}
\usepackage{caption}
\usepackage{calc} 
\usepackage{etoolbox}
\makeatletter
\patchcmd\longtable{\par}{\if@noskipsec\mbox{}\fi\par}{}{}
\makeatother
\IfFileExists{footnotehyper.sty}{\usepackage{footnotehyper}}{\usepackage{footnote}}
\makesavenoteenv{longtable}
\providecommand{\tightlist}{%
  \setlength{\itemsep}{0pt}\setlength{\parskip}{0pt}}
\usepackage{graphicx}
\usepackage{float}
\makeatletter
\newcommand{\keepspace}[1]{\par\penalty0\begingroup\dimen@=#1\relax\dimen@ii\pagegoal\advance\dimen@ii-\pagetotal\ifdim\dimen@>\dimen@ii\ifdim\dimen@ii>\z@\vfil\fi\break\fi\endgroup}
\makeatother
\usepackage{multicol}
\usepackage{etoolbox}
\AtBeginEnvironment{longtable}{\small}
\usepackage{fancyhdr}
\usepackage{caption}
\usepackage{bookmark}
\IfFileExists{xurl.sty}{\usepackage{xurl}}{} 
\makeatletter
\@ifundefined{xmpquote}{}{}
\makeatother
\hypersetup{
  pdftitle={Learning to Read Events for Exact State Tracking with Frozen Language Models},
  colorlinks=true,
  linkcolor={black},
  filecolor={Maroon},
  citecolor={Blue},
  urlcolor={black},
  pdfcreator={LaTeX via pandoc}}

\title{Learning to Read Events for Exact State Tracking with
Frozen Language Models}
\author{Jeonghoon Lee\\[2pt] {\normalsize Attractor Dynamics Inc.}}
\date{18 November 2026}

\begin{document}
\maketitle
\begin{abstract}
Long event streams require identifying each change and
applying the changes in order. We train a selector on a
frozen language model to read each transition. With API-only
access, we ask the model to list the transitions. An
external integer register composes them in order. A
confidence rule can flag uncertain updates or retain several
candidates. We study products of permutations of five items,
a task that fixed-depth, log-precision transformers cannot
solve on all input lengths under the standard conjecture
that \(\mathrm{TC}^0 \ne
\mathrm{NC}^1\). On two-event windows, a 27B model answered
3 of 100 correctly without visible reasoning and 96 of 100
when it generated intermediate steps. A frozen 0.8B model
with a trained selector read every event in 299 of 299
windows of 43,687 events each. The same package passed 298
of 299 new windows on a fresh machine, missing the target of
0.99 for the one-sided 95\% lower confidence bound. A
separate follow-up sized to allow one failure yielded 472 of
473. On 1,000-event room-swap documents, three of ten models
solved all five documents by reasoning alone. All seven
models tested both ways at this length solved all five with
extraction and the register, with no chain of thought
requested. With concurrent extraction, median wall-clock
time fell from 136.1 to 10.4 s for Sonnet 5.5 and from 313.5
to 28.0 s for GPT-6.1 Sol. At fixed per-token rates (section
5.1), mean cost per document fell by 55\% for Sonnet 5.5 and
44\% for GPT-6.1 Sol. A regular expression with an executor
also solved all 15 fixed-format documents. On 500 free-form
issue-tracker comments, a separately adapted 4B reader
identified the recorded value in 68.8\% of cases, compared
with 61.0\% for a frequency baseline fitted on the same
training projects.
\end{abstract}

\thispagestyle{fancy}

\keepspace{8\baselineskip}

\subsection{1 Introduction}\label{introduction}

Many tasks ask a model to keep track of a state that changes
event by event over a long input: which item is in which
slot after a series of moves, the value of a field after a
long run of edits, a participant's status after a long log.
A language model usually reads a single event correctly. The
difficulty is the composition: the state after event \(t\)
depends on every event before it.

For one class of states the difficulty is structural.
Published analyses place fixed-depth transformers that
compute with logarithmic precision in the circuit class
TC\(^0\) (Merrill and Sabharwal, 2023), and state-space
models such as S4 and Mamba with them (Merrill, Petty and
Sabharwal, 2024). The word problem of a non-solvable group
such as \(S_5\), the permutations of five items, is complete
for NC\(^1\) (Barrington, 1989). Under the standard
conjecture that TC\(^0 \neq\) NC\(^1\), such a model cannot
track an \(S_5\) state over arbitrarily long inputs. Trained
models learn shortcuts that hold at training lengths and
fail beyond them, and constant-depth shortcuts exist when
the group is solvable (Liu et al., 2023). This prior result
motivates our choice of \(S_5\). Its asymptotic bound is
specific to the computational model and does not establish
the cause of errors on finite test inputs.

\textbf{Chain of thought crosses this boundary, at a price.}
Generating intermediate steps gives a transformer additional
sequential computation. Under standard conjectures, a
transformer with a number of such steps linear in input
length can recognize every regular language (Merrill and
Sabharwal, 2024), including the word problem for \(S_5\). We
measure the number of reasoning tokens generated at each
input length (section 5.2). On 1,000-event room-swap
documents, three of ten models solved all five by reasoning
alone and five solved at most one (section 5, Figure 1).

\textbf{We split the work at the boundary instead.} We
assign the two operations to different parts of the system.
The model reads local transitions. In the self-hosted
setting it reads all event boundaries in one prefill pass.
An external integer register then composes the transitions
in event order. A confidence rule can apply the leading
prediction, apply it with a provisional flag, or retain
several candidates. We call this combination an exact-state
layer (ESL). Exactness refers to the register calculation
given the extracted transitions.

\keepspace{12\baselineskip}

\textbf{Contributions.}

\begin{enumerate}
\def\labelenumi{\arabic{enumi}.}
\tightlist
\item
  An evaluation of unaided composition on \(S_5\) streams.
  Without visible reasoning, the 0.8B and 27B models
  answered 2 of 299 and 3 of 100 two-event windows
  correctly, respectively (section 3).
\item
  A measured study of the split, reading with the model and
  composing outside it, through two paths: a selector on the
  activations of a frozen, self-hosted model, and per-chunk
  extraction through an API (section 2). The split itself is
  not new (section 9). What we add is a confidence value
  that chooses, per event, among exact, flagged and
  set-valued transitions, with a status that travels with
  the state and decides what may follow.
\item
  Stream-level certificates per host, with the window
  length, sample size and numeric headroom attached to each,
  and an independent re-run of one certificate that did not
  reach its bound (section 4).
\item
  Measured cost and time against reasoning alone, at up to
  1,000 events, on closed and open models (section 5), and
  accuracy on a third-party benchmark (section 6).
\item
  Measured limits: where a parser is enough, where the model
  alone is better, and where the commit policy withheld
  mostly correct writes (sections 5.5, 7, 8 and 10).
\end{enumerate}

\begin{figure}[H]
\centering
\includegraphics[width=0.85\linewidth]{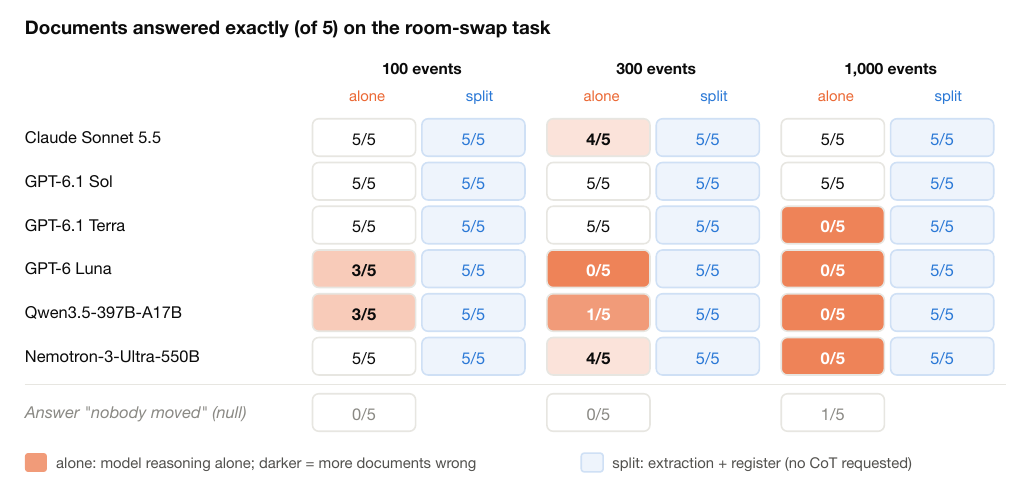}
\caption{Exact final state on the room-swap task, an $S_5$ word problem: documents answered exactly, of five per length. Orange: the model reasoning alone (Sonnet with adaptive thinking; GPT models at effort high; the open models with thinking on, temperature 1.0); darker cells have more documents wrong. Blue: extraction and the register, with no chain of thought requested (thinking off or the lowest effort accepted; the open models at temperature 0; at that lowest setting GPT-6 Luna still produced reasoning tokens, a median of 3,431 per 1,000-event document). Gray row: a null that answers the starting rooms without reading the document; it is right only on the one 1,000-event document that happened, by chance, to end where it started.}
\label{fig:swap}
\end{figure}

\keepspace{8\baselineskip}

\subsection{2 Method}\label{method}

Figure 2 shows how the reader and register process an event
stream.

\subsubsection{2.1 Events, transitions and the
register}\label{events-transitions-and-the-register}

The state lives in a finite algebra; in our main task it is
a permutation of five items, an element of \(S_5\), stored
as an integer index into a \(120 \times 120\) multiplication
table. Each event carries one transition element. The
register composes elements by table look-up, in integers
only. Because the operation is associative, the prefix
products can be computed serially or by a parallel scan with
identical results. No floating-point arithmetic touches the
state.

The model predicts each transition. Given those predictions,
the register computes the resulting state exactly.

\begin{figure}[H]
\centering
\includegraphics[width=0.80\linewidth]{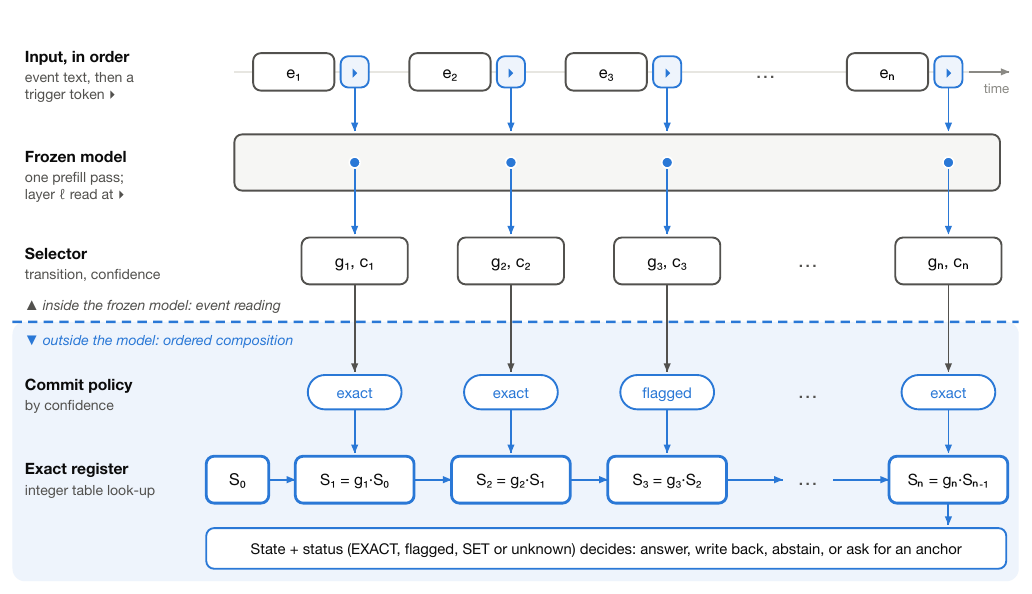}
\caption{The split, drawn along the input from left to right. Each event is followed by a trigger token. Inside the frozen model a selector scores possible transitions from intermediate-layer features aligned to each event boundary, all obtained in one prefill pass. The two highest scores determine the confidence value. Outside the model a commit policy chooses an exact, flagged or set-valued transition, and an exact integer register composes the transitions; the resulting status indicates which downstream actions are permitted. The commit labels shown are an example. The complexity argument in section 1 concerns arbitrary-length products in $S_5$ for fixed-depth, log-precision models.}
\label{fig:split}
\end{figure}

\keepspace{6\baselineskip}

\subsubsection{2.2 Commit policy and
status}\label{commit-policy-and-status}

A confidence value is computed for each read; in the
self-hosted path it is the gap between the selector's two
highest scores. Two thresholds choose one of three
transitions:

\begin{itemize}
\tightlist
\item
  \textbf{exact:} the top-1 element is applied;
\item
  \textbf{flagged:} the top-1 element is applied and the
  state is marked provisional;
\item
  \textbf{candidate set:} a set of elements is applied to
  every candidate state, so the register holds a set of
  possible states, which a later observation (an anchor) can
  narrow.
\end{itemize}

The status that travels with the state (EXACT, flagged, SET,
unknown) tells the downstream system whether it may act. An
event is never skipped while the state stays labelled exact:
in an earlier pilot, silently skipping low-confidence events
reduced final-state exact match from 80.2\% to between 0.5\%
and 1.5\%.

\keepspace{6\baselineskip}

\subsubsection{2.3 Path A: a selector on a frozen,
self-hosted
model}\label{path-a-a-selector-on-a-frozen-self-hosted-model}

The host weights remain fixed. A small selector scores
possible transitions from intermediate-layer features
aligned to each event boundary. In the Qwen3.5-0.8B-Base
package, the features span the event ending at the trigger.
In a 262,140-token window, one prefill pass up to the
selected layer provides features for every event.

Path A requires access to one intermediate layer and the
positions of the event triggers in the token stream. For
each host model, we train the selector on generated event
streams, then evaluate it on fresh windows (section 4).

\textbf{Start of stream.} Before the first real event the
input carries a short preamble of synthetic events with a
known transition: three events whose correct transition is
the identity, each followed by a trigger token. The
selector's outputs at these positions are discarded, and the
register starts at the identity at the first real event.
Without the preamble the first real event was read
unreliably: in one test of 23 windows, 21 read the first
event as the same wrong class. With it, the first event was
read correctly in 23 of 23 windows, and there was no error
in 1,004,801 real events (preregistered). Our working
hypothesis, not a measured cause, is that the first trigger
token has no earlier trigger in its context, so its hidden
state comes from a different distribution than later ones.
Placing priming inputs first and discarding their outputs is
related to attention sinks in streaming language models
(Xiao et al., 2023), register tokens in vision transformers
(Darcet et al., 2024) and burn-in for recurrent agents
(Kapturowski et al., 2019). The automatically fitted
packages used a preamble of eight fixed elements whose
product is the identity.

\keepspace{6\baselineskip}

\subsubsection{2.4 Path B: extraction through an
API}\label{path-b-extraction-through-an-api}

When model activations are unavailable, we divide the
document into chunks and ask the model to extract each
event's change. In the Claude and OpenAI room-swap runs, the
model returns a JSON list of swapped pairs. We record any
discrepancy between the number of pairs and the number of
input lines. In these runs we do not retry the request. For
the open-model, local-4B and StateBench runs, the model
returns numbered lines. If the line count is wrong, we retry
the chunk once. A second count mismatch causes the system to
return no answer for that document. The register applies the
extracted changes in order. We did not request a chain of
thought and used the lowest reasoning setting each provider
accepted. GPT-6 Luna still generated reasoning tokens. One
provider's structured-data mode truncated replies. Numbered
lines with the count check returned complete replies in that
setting (section 10).

\keepspace{8\baselineskip}

\subsection{3 Where unaided composition
fails}\label{where-unaided-composition-fails}

We prompted the models with the \(S_5\) event streams used
for certification and asked for the state after \(n\) events
(Figure 3). Without visible reasoning, Qwen3.5-0.8B-Base
answered 296 of 299 one-event windows and 2 of 299 two-event
windows correctly. Qwen3.8-27B answered 3 of 100 two-event
windows correctly (preregistered). Its responses typically
copied the current event as the new state. When generating
intermediate steps, the 27B model answered 96 of 100
two-event windows and 99 of 100 four-event windows
correctly. In the exploratory tests, added after the first
results, every window was correct at 8, 16 and 32 events,
with about 230 to 250 generated tokens per event at these
lengths. At 64 events, the full trajectory was correct in 23
of 25 windows and the final state in 24. At 128 and 256
events, 6 of 8 and 1 of 4 trajectories were correct,
respectively. No run reached its output limit. The unaided
models received task instructions. The 0.8B model with the
register used a selector trained on event-level labels and
read every event in 299 of 299 windows of 43,687 events
(section 4). These conditions differ in both supervision and
how the state is computed.

\begin{figure}[H]
\centering
\includegraphics[width=0.70\linewidth]{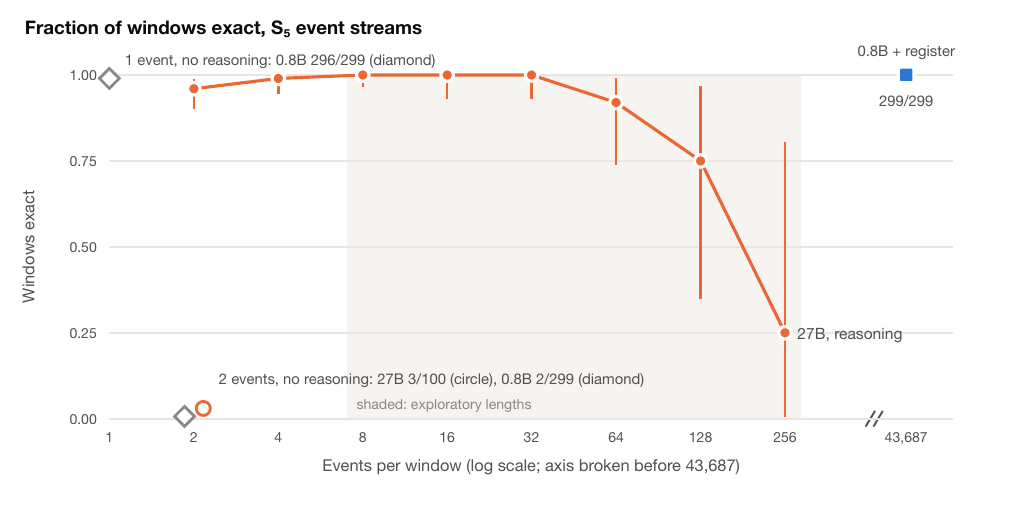}
\caption{Window accuracy on $S_5$ streams by length and inference method. Orange: Qwen3.8-27B with reasoning, fraction of windows exact with two-sided 95\% Clopper--Pearson intervals (100 windows at 2, 4 and 8 events; 50 at 16 and 32; 25 at 64; 8 at 128; 4 at 256). Lengths of 8 and more were added after the first results (shaded). Hollow markers: no visible reasoning. Blue: the frozen 0.8B model with the register, 299 of 299 windows of 43,687 events.}
\label{fig:limit}
\end{figure}

\keepspace{8\baselineskip}

\subsection{4 Certification}\label{certification}

\subsubsection{4.1 What a certificate
states}\label{what-a-certificate-states}

A window is a fixed-length input (262,140 tokens unless
stated) holding about 43,700 real events after the preamble
of section 2.3, whose outputs are discarded. A window passes
only if every event in it is read correctly. A certificate
reports, for one host on one instance and one task, the
one-sided 95\% Clopper-Pearson lower bound on the
probability that a fresh window passes, over \(N\) fresh
windows whose seeds are fixed before the run. We report the
length with every bound. Each window is independently
sampled, with a fresh seed, a separate forward pass and no
cache carried over between windows. Errors within a window
may be dependent. We also report event-level error counts as
descriptive statistics.

At \(N = 299\) a flawless run clears 0.99 by 0.000031
(\(0.05^{1/299} = 0.990031\)), so one failure anywhere
excludes a 0.99 bound (\(298/299\) gives 0.984233);
tolerating one failure at 0.99 needs \(N = 473\)
(\(472/473\) gives 0.990010).

\begin{figure}[H]
\centering
\includegraphics[width=0.70\linewidth]{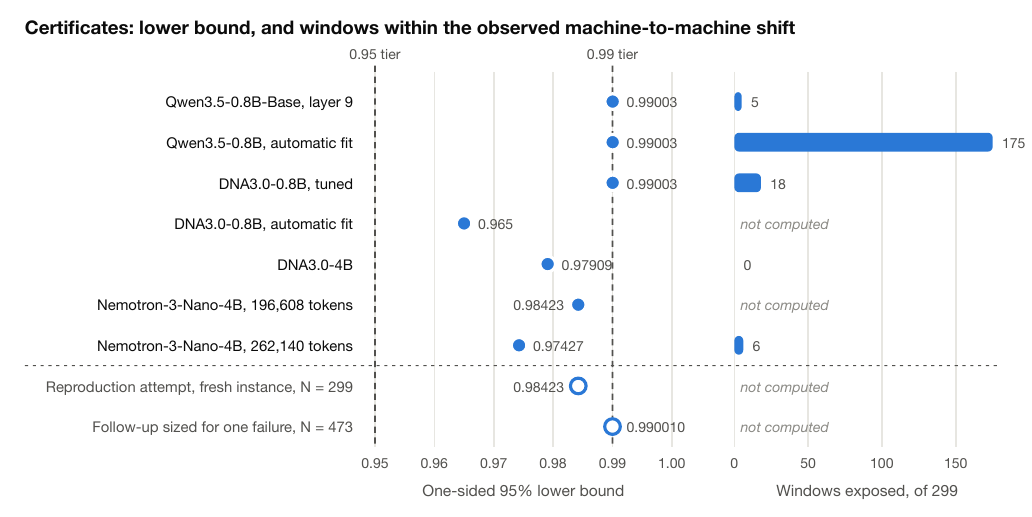}
\caption{Window certificates: one-sided 95\% Clopper--Pearson lower bounds (262,140-token windows unless stated), and, at $N = 299$, the windows whose tightest correct event has a gap below 46.9, the largest change in a gap observed between two machines (a sensitivity count, not a bound). Hollow markers: re-runs of the first package on fresh instances, an independent reproduction attempt (N = 299) and a follow-up sample sized to tolerate one failure (N = 473).}
\label{fig:certs}
\end{figure}

\keepspace{14\baselineskip}

\subsubsection{4.2 Certificates
issued}\label{certificates-issued}

{\def\LTcaptype{none} 
\begin{longtable}[]{@{}
  >{\raggedright\arraybackslash}p{(\linewidth - 12\tabcolsep) * \real{0.2667}}
  >{\raggedright\arraybackslash}p{(\linewidth - 12\tabcolsep) * \real{0.1389}}
  >{\raggedright\arraybackslash}p{(\linewidth - 12\tabcolsep) * \real{0.0167}}
  >{\raggedright\arraybackslash}p{(\linewidth - 12\tabcolsep) * \real{0.0833}}
  >{\raggedright\arraybackslash}p{(\linewidth - 12\tabcolsep) * \real{0.0611}}
  >{\raggedright\arraybackslash}p{(\linewidth - 12\tabcolsep) * \real{0.1778}}
  >{\raggedright\arraybackslash}p{(\linewidth - 12\tabcolsep) * \real{0.2556}}@{}}
\toprule\noalign{}
\begin{minipage}[b]{\linewidth}\raggedright
Host (revision)
\end{minipage} & \begin{minipage}[b]{\linewidth}\raggedright
Window, tokens
\end{minipage} & \begin{minipage}[b]{\linewidth}\raggedright
\(N\)
\end{minipage} & \begin{minipage}[b]{\linewidth}\raggedright
Failing windows
\end{minipage} & \begin{minipage}[b]{\linewidth}\raggedright
Lower bound
\end{minipage} & \begin{minipage}[b]{\linewidth}\raggedright
Smallest correct gap (\(N = 299\))
\end{minipage} & \begin{minipage}[b]{\linewidth}\raggedright
Windows with tightest correct event below 46.9
\end{minipage} \\
\midrule\noalign{}
\endhead
\bottomrule\noalign{}
\endlastfoot
Qwen3.5-0.8B-Base (layer 9) & 262,140 (262,144 nominal) &
299 & 0 & 0.99003 & 10.46 & 5 \\
Qwen3.5-0.8B (rev 2fc06364), automatic fit & 262,140 & 299 &
0 & 0.99003 & 8.97 & 175 \\
dnotitia DNA3.0-0.8B (rev ec36b145, layer 6), tuned &
262,140 & 299 & 0 & 0.99003 & 23.17 & 18 \\
dnotitia DNA3.0-0.8B, automatic fit & 262,140 & 299 & 5 &
0.965 & & \\
dnotitia DNA3.0-4B & 262,140 & 299 & 2 & 0.97909 & 56.74 &
0 \\
NVIDIA Nemotron-3-Nano-4B (layer 11) & 196,608 & 299 & 1 &
0.98423 & & \\
NVIDIA Nemotron-3-Nano-4B (layer 11) & 262,140 & 299 & 3 &
0.97427 & 16.28 & 6 \\
\end{longtable}
}

The gap is the difference between the selector's top two
scores at an event. 46.9 is the largest change in a gap that
we have observed between two machines running the same
frozen inputs. A correct event whose gap is below it could,
in principle, be read differently on another machine; the
last column counts the windows that hold at least one such
event. It is a sensitivity count, not a bound: another
machine could shift a gap by more than we have seen.

Three packages had the same lower bound of 0.99003 (Figure
4). Of 299 windows, 5 for the original Qwen package, 18 for
the tuned DNA3.0 package and 175 for the automatically
fitted Qwen package contained a correct prediction with a
gap below 46.9. The smallest correct gap for the original
Qwen package was 10.46 at \(N=299\) and 6.52 in a separate
sample of 473 (section 4.3). We give the sample size with
each minimum.

A small correct gap and a prediction error measure different
outcomes. DNA3.0-4B made two errors even though none of its
correct predictions had a gap below 46.9. The automatically
fitted Qwen package made no errors in its sample despite 175
windows containing a correct prediction below that
threshold.

In the fourth Nemotron-3-Nano-4B run at 262,140 tokens, 30
of 299 windows failed, with errors concentrated near the end
of the window. With the selection set held fixed, we
increased the number of training examples per late position.
Late-band errors fell from 30 to 2 per 510,393 events. The
fifth run passed 296 of 299 windows, with a one-sided 95\%
lower bound of 0.97427.

\keepspace{6\baselineskip}

\subsubsection{4.3 A fresh-instance re-run and a separate
follow-up}\label{a-fresh-instance-re-run-and-a-separate-follow-up}

We tested the first package in the table on a fresh instance
with 299 new windows. The seeds were fixed in advance and we
verified the source files byte for byte. The package passed
298 of 299 windows (one-sided 95\% lower bound 0.984233),
below the 0.99 target. The only misread occurred at event
278. The wrong top prediction led the runner-up by 38.0. The
register state was incorrect at every position from event
278 through event 43,687, a total of 43,410 positions. Gaps
also varied between runs on the same new instance, and this
event was never scored on the original instance. The cause
of the misread remains unknown.

We then chose a sample size of 473 so that one failure would
still meet the 0.99 target. The seeds were hashed before the
instance existed. In this separate run, 472 of 473 windows
passed (lower bound 0.990010). We report the samples
separately. With the fitted selector held fixed, we
evaluated 473 fresh windows and three control windows in 54
minutes on a single RTX 5090. The provider billed US\$0.63
for the run, or about US\$0.03 per million events in the 473
fresh windows.

\keepspace{6\baselineskip}

\subsubsection{4.4 Errors and the candidate
set}\label{errors-and-the-candidate-set}

In the 4B runs that record each read's runner-up, the true
transition was the selector's second choice in all 34 errors
of DNA3.0-4B, Nemotron-3-Nano-4B at 196,608 tokens and
Nemotron's fourth attempt at 262,140 (31 of them from that
attempt, which did not certify). In a replay of recorded
outputs under a protocol frozen beforehand, all 24 errors in
the frozen set fell at or below the threshold and were
carried as candidate sets: no wrong state was released as
EXACT, where a register that always commits its top choice
carried a wrong state at 89,964 positions after the same 24
errors, and the true state stayed in the set in every
window. The price is availability: without an anchor, 74\%
to 99.7\% of positions in long windows were not EXACT. A
25th error, found after the freeze, had a gap of 260.7,
above that run's preregistered threshold of 188.3, and was
committed: a confident misread. It released no wrong EXACT
only because earlier uncertain reads had already put that
window in a candidate set.

\keepspace{8\baselineskip}

\subsection{\texorpdfstring{5 A 1,000-event task in \(S_5\):
reasoning against the
split}{5 A 1,000-event task in S\_5: reasoning against the split}}\label{a-1000-event-task-in-s_5-reasoning-against-the-split}

\subsubsection{5.1 Setup}\label{setup}

Each document lists five people, each in a room, and then
\(L\) lines of the form ``\(k\). X and Y swapped rooms'',
followed by a question asking for every person's room. A
swap of two people's rooms is a transposition, and
transpositions generate \(S_5\), so the task is an \(S_5\)
word problem. The documents are written in Korean. Five
documents per length, \(L \in \{100, 300, 1000\}\),
generated from fixed seeds, with the truth computed by a
serial executor. A document counts as exact only if all five
rooms are right. One 1,000-event document ends where it
started, so the answer ``nobody moved'' scores five of five
on it; that null averages 1.4 people right at 1,000 events.
Runs took place from 12 to 18 November 2026 on the model
versions then served. Costs are computed, not billed: the
input and output token counts that each API reported,
multiplied by a fixed rate per million tokens (US\$2 input
and US\$10 output for Sonnet 5.5 and GPT-6.1 Sol; the rate
for every model is in the released run scripts). Output
counts include reasoning tokens; no caching discount is
applied.

Reasoning arms received the document and one line asking
them not to run code or use tools. Extraction arms (Path B)
received 50-line chunks and returned the swapped pair of
each line.

\keepspace{6\baselineskip}

\subsubsection{5.2 Reasoning alone}\label{reasoning-alone}

At 1,000 events, Claude Sonnet 5.5 and Claude Opus 5.5
(adaptive thinking) and GPT-6.1 Sol (effort high) answered
five of five; GPT-6.2 Sol four (three of the four documents
that move); Gemini 4 carbon-high four and Grok 4.7 xhigh
one, both through a coding client in a read-only mode;
GPT-6.1 Terra, GPT-6 Luna, Qwen3.5-397B-A17B and
Nemotron-3-Ultra-550B-A55B none. Terra, Luna and Qwen-397B
averaged fewer people right than the ``nobody moved'' null
(1.2, 0.8 and 0.6 against 1.4). Qwen-397B returned the
initial room assignments. Nemotron-Ultra reached its
32,768-token output limit on every 1,000-event document
without producing a final answer, both at temperature 1.0
and under greedy decoding. Sonnet's median reasoning-token
counts were 3,092, 9,037 and 27,661 at 100, 300 and 1,000
events, respectively. Sol's were 2,389, 8,169 and 15,534. At
1,000 events, Opus used 24,906 reasoning tokens (161.6 s and
US\$0.58 per document).

\keepspace{6\baselineskip}

\subsubsection{5.3 Accuracy, time and cost with extraction
and the
register}\label{accuracy-time-and-cost-with-extraction-and-the-register}

Seven models were tested with extraction at all three
lengths: Sonnet, GPT-6.1 Sol, Terra, Luna, gpt-4.1-mini,
Qwen-397B and Nemotron-Ultra. Each extracted every event
correctly and solved all five documents at every length.
GPT-6.2 Sol, tested only at 1,000 events, also read every
event correctly and solved all five documents. No chain of
thought was requested. Only Luna generated reasoning tokens
during extraction (median 3,431 per 1,000-event document).

For Sonnet and GPT-6.1 Sol, both methods solved all five
1,000-event documents. With concurrent extraction, median
wall-clock time fell from 136.1 to 10.4 s for Sonnet and
from 313.5 to 28.0 s for Sol (speedups of 13.0 and 11.2,
calculated from unrounded medians). Mean cost per document
fell from US\$0.333 to US\$0.148 for Sonnet and from
US\$0.203 to US\$0.114 for Sol. Summing the chunk-call
durations within each document gives estimated serial times
of 83.1 and 150.6 s, respectively. GPT-6.2 Sol used fewer
reasoning tokens alone than GPT-6.1 Sol (11,289 versus
15,534). Extraction raised its score from four to five
correct documents and reduced median wall-clock time by a
factor of 17.0. Mean cost per document was US\$0.138 alone
and US\$0.129 with extraction (Figure 5).

\keepspace{6\baselineskip}

\subsubsection{5.4 Extraction with two 4B models without
fine-tuning}\label{extraction-with-two-4b-models-without-fine-tuning}

Served as released, with no fine-tuning, dnotitia DNA3.0-4B
and NVIDIA Nemotron-3-Nano-4B at first failed at 1,000
events: each rewrote the line numbers of one chunk at a
fixed position, and the count check rejected it. Stating the
line range in each chunk's prompt (``lines \(a\) to \(b\);
keep the original numbers'') removed the failure. Both were
then exact on 5 of 5 at 100 and 300 events and 3 of 5 at
1,000 (DNA3.0-4B abstained on one document and answered one
wrong; Nemotron answered two wrong), misreading 1 of 4,950
and 2 of 5,000 events, in about 12 s per 1,000-event
document on one 20 to 24 GB GPU. Each wrong answer came from
a single misread pair: one misread in 1,000 events breaks an
exact state. A shuffled-document control confirms that the
models read the lines in order.

\begin{figure}[H]
\centering
\includegraphics[width=0.70\linewidth]{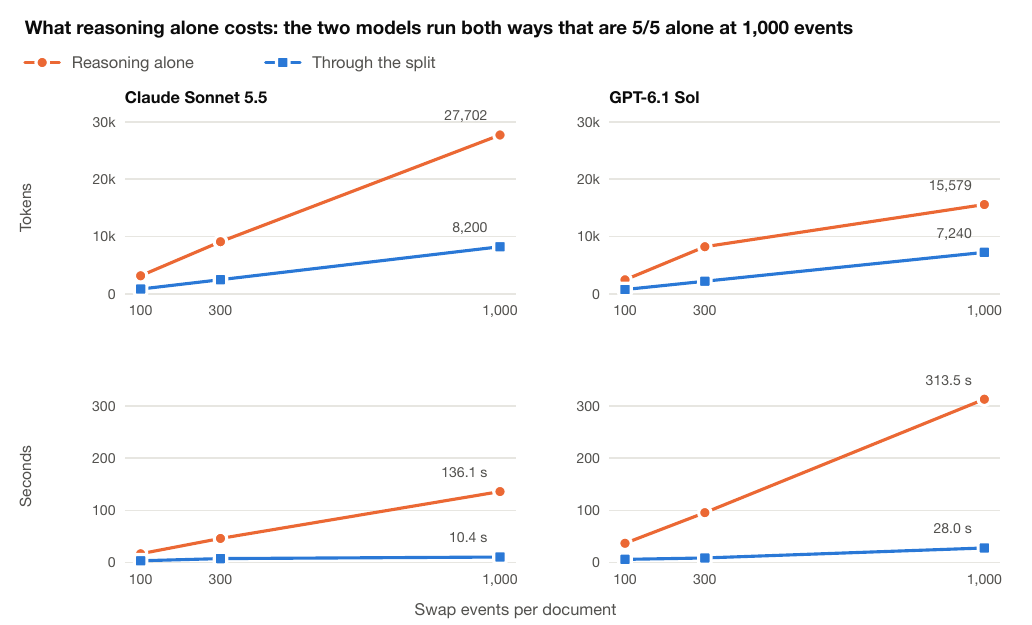}
\caption{Generated tokens and wall time per room-swap document for the two models run both ways at every length that answer 1,000-event documents correctly when reasoning alone. All but about 45 of each document's tokens on the reasoning arm are thinking tokens; the split generates about 8 output tokens per event and none of them thinking. Extraction requests ran concurrently. Summing the chunk durations within each 1,000-event document gives estimated serial times of 83.1~s (Sonnet) and 150.6~s (Sol).}
\label{fig:cost}
\end{figure}

\keepspace{6\baselineskip}

\subsubsection{5.5 A fixed-format parser
baseline}\label{a-fixed-format-parser-baseline}

A regular expression identified the swapped pair in each
line. Combined with the executor, it solved all 15
documents, with a median runtime of 0.26 milliseconds at
1,000 events. Section 7 evaluates a learned reader on
changes described in free-form comments.

\subsection{6 A third-party
benchmark}\label{a-third-party-benchmark}

StateBench (Petty, Merrill and Sabharwal, 2026) is a
hierarchy of four tiers of state tracking, rendered in
several surface formats: r-trivial (immutable assignment),
aperiodic (mutable assignment), solvable (self-referential
assignment, including additions modulo 100) and general
finite state tracking, which the release names periodic. In
the periodic tier the values of 2 to 16 variables are
swapped and reset, so its items include non-solvable
permutation groups. \textbf{Only the periodic tier bears on
the TC\(^0\) argument.} The other three hold no non-solvable
group, so the argument does not apply to them; there the
benchmark measures reading, cost and transfer. Items were
drawn by the release's own sampling rule (seed 0) from the
public data, not taken from the authors' frontier set.

\keepspace{6\baselineskip}

\subsubsection{6.1 StateBench results on OLMo and
Qwen}\label{statebench-results-on-olmo-and-qwen}

On OLMo-Hybrid-7B, we matched the training items to the test
items in length and filler density. The reader passed the
prespecified accuracy threshold on integer-code and
spreadsheet items, with exact final states on 576 of 576 and
575 of 576 items, respectively. Both formats were
represented in training. In a separate test, the
Qwen3.5-0.8B-Base selector was trained on two formats and
evaluated on a third, the spreadsheet format. It produced no
exact final states on that format, and its five-choice
accuracy was at chance level.

For comparison, we fine-tuned OLMo-3-7B with low-rank
adaptation on the same 26,400 items, using final-state
labels. The selector was trained on per-statement labels.
The register system led in 5 of 6 periodic cells at 4k to
16k tokens. The fine-tuned model matched or exceeded it in
19 of 24 solvable and aperiodic cells. The differences in
host and supervision limit what can be attributed to the
register.

\keepspace{6\baselineskip}

\subsubsection{6.2 API results under output-token
limits}\label{api-results-under-output-token-limits}

We used the benchmark authors' prompt and parser with a
16,384-token output cap at the provider's default effort. A
capped reply usually has no answer line and is scored wrong
(one Opus reply, periodic 32k, had written it and was scored
correct), so we also give the accuracy among the items that
finished.

{\def\LTcaptype{none} 
\begin{longtable}[]{@{}
  >{\raggedright\arraybackslash}p{(\linewidth - 12\tabcolsep) * \real{0.2105}}
  >{\raggedright\arraybackslash}p{(\linewidth - 12\tabcolsep) * \real{0.0439}}
  >{\raggedright\arraybackslash}p{(\linewidth - 12\tabcolsep) * \real{0.0702}}
  >{\raggedright\arraybackslash}p{(\linewidth - 12\tabcolsep) * \real{0.1842}}
  >{\raggedright\arraybackslash}p{(\linewidth - 12\tabcolsep) * \real{0.2632}}
  >{\raggedright\arraybackslash}p{(\linewidth - 12\tabcolsep) * \real{0.1491}}
  >{\raggedright\arraybackslash}p{(\linewidth - 12\tabcolsep) * \real{0.0789}}@{}}
\toprule\noalign{}
\begin{minipage}[b]{\linewidth}\raggedright
Model, cell (n = 30)
\end{minipage} & \begin{minipage}[b]{\linewidth}\raggedright
Alone
\end{minipage} & \begin{minipage}[b]{\linewidth}\raggedright
Cap hits
\end{minipage} & \begin{minipage}[b]{\linewidth}\raggedright
Alone, among finished
\end{minipage} & \begin{minipage}[b]{\linewidth}\raggedright
Alone, cap 65,536, effort high
\end{minipage} & \begin{minipage}[b]{\linewidth}\raggedright
Through the split
\end{minipage} & \begin{minipage}[b]{\linewidth}\raggedright
Code tool
\end{minipage} \\
\midrule\noalign{}
\endhead
\bottomrule\noalign{}
\endlastfoot
Opus 5.5, solvable 32k & 17/30 & 13 & \textbf{17/17} & &
29/30 & 30/30 \\
Opus 5.5, periodic 32k & 4/30 & 24 & 3/6 & & 30/30 &
30/30 \\
GPT-6 Sol, solvable 32k & 1/30 & 2 & 1/28 & 4/30 & 30/30 &
25/30 \\
GPT-6 Sol, periodic 32k & 0/30 & 4 & 0/26 & 1/30 & 30/30 &
27/30 \\
GPT-6 Luna, solvable 16k & 1/30 & 2 & 1/28 & 2/30 & 30/30 &
25/30 \\
GPT-6 Luna, periodic 16k & 2/30 & 0 & 2/30 & 2/30 & 29/30 &
24/30 \\
\end{longtable}
}

\textbf{Opus on solvable 32k is not evidence of lost state.}
It was right on all 17 items it finished and wrong on all 13
it did not, so that cell measures the cap and is excluded
from the claim. The claim rests on Sol and Luna, which hit
the cap on 0 to 4 of 30 items and still failed, and which a
fourfold cap with high effort moved from at most 2 to at
most 4 of 30. Where events can be parsed, a program written
by the model also reaches high counts (the last column); the
split is not the only route to them on this benchmark.

\keepspace{8\baselineskip}

\subsection{7 Reading prose: issue-tracker
records}\label{reading-prose-issue-tracker-records}

We selected anonymised public issue-tracker feeds from
Apache projects (CC BY 4.0) using a rule fixed in advance.
Each log query can be answered from the value at creation or
the latest update to the field. We used the comments to test
whether a reader can identify the recorded change from
prose.

A comment such as ``moving this to the next release'' can
announce a change without stating the new field value
uniquely. The unadapted reader fell below the prespecified
threshold. We adapted Qwen3.5-4B with low-rank training on
61,666 items from projects excluded from testing. On 500
free-form change comments, it predicted the recorded value
with 68.8\% accuracy. A frequency baseline fitted on the
same training projects reached 61.0\%, a difference of 7.8
percentage points (project-clustered 95\% interval, 2.8 to
12.6 points). Accuracy was 56.8\% for a model trained on the
creation line alone, 4.6\% for a hand-written parser, and
9.6\% for GPT-6.1 Terra reading the whole record.
Withholding the comment reduced the adapted reader's
accuracy by 17.4 percentage points. On 100 comments with an
explicitly worded change, accuracy was 92\% for the reader
and 85\% for the parser. At a strict threshold the reader
committed to a value on 22 of the 500 free-form items
(4.4\%), with no errors among those 22. The reader
evaluation was preregistered. The parser and Terra
comparisons were exploratory. We tested one training run on
one host family.

\keepspace{8\baselineskip}

\subsection{8 Where the split does not
help}\label{where-the-split-does-not-help}

\begin{itemize}
\tightlist
\item
  \textbf{Free conversation.} On 150 LongMemEval state
  questions (Wu et al., 2025), GPT-6.1 Terra scored 0.947
  alone and 0.860 with the register.
\item
  \textbf{The commit policy on dialogue state.} On MultiWOZ
  2.4 (Ye et al., 2022), we replayed GPT-6.1 Terra's outputs
  and withheld uncertain operations by token probability.
  Incorrect writes fell from 575 to 474. Withholding the
  same number at random yielded 497.5 to 524.0 incorrect
  writes in the reported comparison. In the live run, the
  system wrote 509 incorrect changes. Of the 1,288 changes
  it withheld, 1,162 would have been correct. Joint goal
  accuracy fell from 0.6175 to 0.5042.
\end{itemize}

\keepspace{8\baselineskip}

\subsection{9 Related work}\label{related-work}

\textbf{Limits of fixed-depth models.} Log-precision
transformers are in TC\(^0\) (Merrill and Sabharwal, 2023);
the cited analysis also bounds S4 and Mamba under its
precision assumptions (Merrill, Petty and Sabharwal, 2024);
the word problem of a non-solvable group is
NC\(^1\)-complete (Barrington, 1989); trained transformers
learn shortcuts to automata that fail beyond training
lengths, with constant-depth shortcuts for solvable groups
(Liu et al., 2023); and a chain of thought linear in the
input length lets a transformer recognize every regular
language (Merrill and Sabharwal, 2024). Section 5 compares
the accuracy, time and cost of reasoning alone with those of
extraction and external composition.

\textbf{Discrete computation attached to a model.} PD-SSM
(Terzić et al., 2025) composes hard-chosen transition
matrices inside the layer. Dhanraj and Eliasmith (2025)
apply a symbolic rule to one layer's encoded state behind a
fixed similarity gate that decides only whether to
intervene. IGC (Dietz and Klakow, 2025) maps hidden
activations to an exact calculator, one operation per
sample, with no accumulated state. Frozen Tape Modules
(Granitzer et al., 2026), the closest, read a tape of
discrete symbols from a frozen small model, rewrite it with
a learned network that is then frozen, and answer all 750
items of BIG-Bench Hard's shuffled-objects task. Every
selection is an argmax at evaluation (a soft relaxation is
used in training), and a layer abstains by writing an empty
tape, which closes its gate and leaves the base model
unchanged: that decides whether the module acts, not which
transition enters a state. None of these uses a per-event
confidence value to choose among transitions, and none
carries a status with the state.

\textbf{Abstention and sets of possible states.} Conformal
abstention (WO 2025/165950 A1) and uncertainty-based
abstention (WO 2025/245362 A1; Farquhar et al., 2024) decide
whether to answer one prompt and accumulate nothing. Sets or
distributions over states are standard in discrete-event
diagnosis under uncertain observations (Su, Zanella and
Grastien, 2016) and in multi-object tracking over
permutations (Huang, Guestrin and Guibas, 2009, approximate
and with no commit-or-defer gate). What is new here is the
combination: a learned reader's confidence choosing among
exact, flagged and set-valued transitions, exact composition
over long inputs, and certification per host.

\keepspace{8\baselineskip}

\subsection{10 Limits, and who should not use
this}\label{limits-and-who-should-not-use-this}

\begin{itemize}
\tightlist
\item
  We evaluated composition on generated permutation streams,
  fixed-format room-swap documents and StateBench. The
  issue-tracker study tests the reading of individual
  changes.
\item
  Each bound applies to the tested model, instance, task and
  window length. A fresh instance missed the original 0.99
  target (section 4.3). We recommend a fresh evaluation on
  the deployment configuration. The certificate files do not
  record the device or the numeric precision, and the
  fitting pipeline did not verify either on the running
  host.
\item
  The register computes the composition of the transitions
  it receives. In the prespecified replay, retaining
  candidate sets yielded no incorrect EXACT outputs, at the
  cost of availability. A subsequent high-confidence misread
  was committed in a window that was already in the
  candidate-set state (section 4.4).
\item
  One small commercial API model reached at best one of five
  at 1,000 events through the split. Without a count check,
  27 of its 140 chunk replies across all lengths had the
  wrong number of lines, and 7 of 5,650 pairs were misread.
\item
  Not for events a parser can read, or for state in free
  conversation (sections 5.5 and 8).
\end{itemize}

\keepspace{8\baselineskip}

\subsection{11 Data and code
availability}\label{data-and-code-availability}

We have prepared a verification package for
\url{https://github.com/jeonghoon-ad/learning-to-read-events}.
It includes document generators, item identifiers and seeds,
scoring scripts, the register, per-window certification
records, and API transcripts except those of the unnamed
model in section 10. StateBench items are provided as
identifiers with regeneration code. The package supports
rescoring the recorded runs. Selector checkpoints, training
recipes and the fitting pipeline are withheld. No patent
licence is granted.

\textbf{Use of AI tools.} The author conceived the method,
defined the research questions, selected the experiments,
and determined which claims the evidence supports. The
author used AI assistants, including Anthropic Claude
(through Claude Code) and OpenAI Codex, to write and run
much of the experimental code, check reported results
against run logs, and draft, critique, and revise the
manuscript. The author takes full responsibility for the
paper.

\footnotesize

\subsection{References}\label{references}

\begin{multicols}{2}

\begin{itemize}
\tightlist
\item
  Barrington, D. A. M. (1989). Bounded-width polynomial-size
  branching programs recognize exactly those languages in
  NC\(^1\). \emph{Journal of Computer and System Sciences}
  38(1), 150--164. doi:10.1016/0022-0000(89)90037-8.
\item
  Darcet, T., Oquab, M., Mairal, J. and Bojanowski, P.
  (2024). Vision transformers need registers. \emph{ICLR
  2024}. arXiv 2309.16588.
\item
  Dhanraj, V. and Eliasmith, C. (2025). Improving rule-based
  reasoning in LLMs using neurosymbolic representations.
  arXiv 2502.01657, v3.
\item
  Dietz, F. and Klakow, D. (2025). IGC: integrating a gated
  calculator into an LLM to solve arithmetic tasks reliably
  and efficiently. arXiv 2501.00684.
\item
  Farquhar, S., Kossen, J., Kuhn, L. and Gal, Y. (2024).
  Detecting hallucinations in large language models using
  semantic entropy. \emph{Nature} 630(8017), 625--630.
  doi:10.1038/s41586-024-07421-0.
\item
  Granitzer, M., Zerhoudi, S., Al-Maamari, M., Dang, H. D.
  and Ghosh Dastidar, K. (2026). A tool in the forward pass:
  Frozen Tape Modules as layers of a small language model.
  arXiv 2611.00783.
\item
  Huang, J., Guestrin, C. and Guibas, L. (2009). Fourier
  theoretic probabilistic inference over permutations.
  \emph{JMLR} 10.
\item
  Kapturowski, S. et al.~(2019). Recurrent experience replay
  in distributed reinforcement learning. \emph{ICLR 2019}.
\item
  Liu, B., Ash, J. T., Goel, S., Krishnamurthy, A. and
  Zhang, C. (2023). Transformers learn shortcuts to
  automata. \emph{ICLR 2023}. arXiv 2210.10749.
\item
  Merrill, W. and Sabharwal, A. (2023). The parallelism
  tradeoff: limitations of log-precision transformers.
  \emph{TACL} 11, 531--545. doi:10.1162/tacl\_a\_00562.
\item
  Merrill, W. and Sabharwal, A. (2024). The expressive power
  of transformers with chain of thought. \emph{ICLR 2024}.
  arXiv 2310.07923.
\item
  Merrill, W., Petty, J. and Sabharwal, A. (2024). The
  illusion of state in state-space models. \emph{ICML 2024},
  PMLR 235, 35492--35506. arXiv 2404.08819.
\item
  Petty, J., Merrill, W. and Sabharwal, A. (2026). A
  fine-grained hierarchy and benchmark for state tracking in
  language models. arXiv 2611.03094.
\item
  Su, X., Zanella, M. and Grastien, A. (2016).
  Diagnosability of discrete-event systems with uncertain
  observations. \emph{IJCAI 2016}, 1265--1271.
\item
  Terzić, A., Menet, N., Hersche, M., Hofmann, T. and
  Rahimi, A. (2025). Structured sparse transition matrices
  to enable state tracking in state-space models (PD-SSM).
  arXiv 2509.22284.
\item
  Wu, D., Wang, H., Yu, W., Zhang, Y., Chang, K.-W. and Yu,
  D. (2025). LongMemEval: benchmarking chat assistants on
  long-term interactive memory. \emph{ICLR 2025}. arXiv
  2410.10813.
\item
  Xiao, G., Tian, Y., Chen, B., Han, S. and Lewis, M.
  (2023). Efficient streaming language models with attention
  sinks. arXiv 2309.17453.
\item
  Ye, F., Manotumruksa, J. and Yilmaz, E. (2022). MultiWOZ
  2.4: a multi-domain task-oriented dialogue dataset with
  essential annotation corrections to improve state tracking
  evaluation. \emph{SIGDIAL 2022}, 351--360.
\item
  WO 2025/165950 A1. Conformal abstention for neural
  networks.
\item
  WO 2025/245362 A1. Detecting neural network hallucinations
  using in-context learning.
\end{itemize}

\end{multicols}

\end{document}